\documentclass[runningheads]{llncs}
\usepackage{esvect} 
\usepackage[T1]{fontenc}
\usepackage{booktabs}
\usepackage{float}
\usepackage[table]{xcolor}
\usepackage{graphicx,verbatim}
\usepackage{amsmath}
\usepackage{amssymb}
\begin{document}
\title{Federated Multi-Task Learning for Bladder Tumor Segmentation and MIBC Classification Using a Hybrid CNN-Transformer Architecture}
%

\author{Malhar Udmale\inst{1}\textsuperscript{*} \and
Divyanshu Dwivedi\inst{2}\textsuperscript{*} \and
Aarohi Dhand\inst{2}\textsuperscript{*} \and Sachin Dudda Nagaraju\inst{3} \and Mayank Rai\inst{2} \and Bagesh Kumar\inst{2}}
\authorrunning{Malhar et al.}
%
\institute{Indian Institute of Information Technology Allahabad, Prayagraj, India \\
\email{iib2024020@iiita.ac.in} \and
Manipal University Jaipur, Jaipur, India
\\
\email{\{divyanshu.2427010047, aarohi.2430040138, mayank.2430040139\}@muj.manipal.edu,
bagesh.kumar@jaipur.manipal.edu}
 \and
Norwegian University of Science and Technology, Trondheim, Norway\\
\email{sachin.d.nagaraju@ntnu.no}
}

\footnotetext{* These authors contributed equally to this work.}

  
\maketitle              
\begin{abstract}
Accurate bladder tumor segmentation and assessment of muscle invasion from
T2-weighted MRI are important for treatment planning, but developing robust
models across institutions is challenging because patient data cannot be
centrally pooled and imaging characteristics vary across scanners and acquisition
protocols. We propose a federated multi-task learning framework for joint bladder
tumor segmentation and MIBC/NMIBC classification across four clinical centers.
The proposed Swin Hybrid model combines a ResNet-34 branch for local texture
and boundary information with a Swin-Tiny Transformer for global anatomical
context. A segmentation-guided classification mechanism further uses tumor
localization information to support MIBC prediction. We also investigate several
augmentation strategies under both centralized and federated training to improve
robustness to multi-center variability. Experiments on the FedBCa dataset show
that the Swin Hybrid provides the best overall balance between segmentation and
classification among the evaluated architectures. Under federated training,
Geo+Elastic augmentation achieved a DSC of 0.8100 and a patient-level AUC of
0.8931, yielding the highest combined score of 0.8474. These results demonstrate
that joint segmentation and classification can be effectively performed across
multiple institutions using federated training without centralizing patient data.
\end{abstract}

\section{Introduction}

Bladder cancer is one of the most common urological malignancies, and accurate staging is essential for determining appropriate treatment \cite{bray2024global}. In particular, distinguishing non-muscle-invasive bladder cancer (NMIBC) from muscle-invasive bladder cancer (MIBC) directly influences clinical management, as MIBC typically requires more aggressive treatment than localized therapies for NMIBC \cite{panebianco2018multiparametric}. In addition to staging, accurate tumor segmentation provides valuable anatomical information for treatment planning, disease monitoring, and quantitative assessment \cite{cha2016bladder}. T2-weighted Magnetic Resonance Imaging (MRI) is widely used for bladder cancer evaluation because of its excellent soft-tissue contrast, but manual tumor delineation and staging remain labor-intensive and are subject to inter-observer variability \cite{panebianco2018multiparametric}. These challenges have motivated the development of automated deep learning methods for the analysis of bladder cancer.

Deep learning has achieved promising performance in medical
image segmentation and classification, but its success
depends on access to diverse datasets of multiple centers
\cite{litjens2017survey}. In clinical practice, patient
data are distributed across hospitals and cannot be
centrally shared due to privacy regulations
\cite{kaissis2020secure}. Federated learning (FL) addresses
this limitation by enabling collaborative model training
without transferring patient data between institutions
\cite{fedavg}. However, multi-center MRI datasets introduce
substantial scanner heterogeneity due to differences in
scanner hardware and acquisition protocols \cite{liu2021feddg}.
These non-IID data distributions create domain shifts that
reduce the robustness and generalizability of federated
models across clinical sites \cite{zhao2018federated}.
Despite these challenges, few studies have explored federated
multi-task learning that jointly addresses tumor segmentation
and staging classification under scanner heterogeneity in
bladder cancer MRI. Existing approaches commonly treat
segmentation and classification as independent tasks
\cite{dolz2018hyperdense}, missing the complementary
anatomical and pathological information shared between them.
Jointly learning both tasks enables the network to learn
shared feature representations that improve diagnostic
performance while reducing overfitting \cite{zhang2021survey}.
Furthermore, a hybrid architecture combining convolutional
neural networks with vision transformers can capture both
fine local boundary information and long-range anatomical
context, making it well suited to heterogeneous multi-center
MRI data \cite{wang2022mixed}.

To address these challenges, we propose a privacy-preserving
federated multi-task learning framework for joint bladder tumor segmentation and MIBC classification using the multi-center FedBCa dataset \cite{cao2024multicenter}. Motivated by recent work on
augmentation-driven domain generalization in federated
medical imaging \cite{nagaraju2025fedgin}, we conduct a systematic study of augmentation strategies designed to harmonize appearance variability across heterogeneous MRI scanners, identifying the combination that best improves cross-center robustness under FL.

The main contributions of this work are as follows:\begin{itemize}
    \item \textbf{Federated Multi-Task Framework:} We develop 
    a federated learning framework using FedAvg that jointly 
    optimizes bladder tumor segmentation and MIBC 
    classification across four heterogeneous clinical centers 
    without sharing patient data.

    \item \textbf{Swin Hybrid Architecture:} We propose a 
    dual-stream encoder combining a pretrained ResNet-34 CNN 
    branch for local texture and boundary features with a 
    Swin-Tiny Transformer branch for global anatomical 
    context, connected via segmentation-guided attention for 
    classification.

    \item \textbf{Augmentation Study for MRI Scanner 
    Heterogeneity:} We systematically evaluate 
    six augmentation strategies under both centralized and 
    federated training, identifying Geo+Elastic deformation as the optimal strategy, achieving DSC of 0.8100 and AUC of 0.8931 under FL.
\end{itemize}

\section{Related Works}

Medical image segmentation and classification are often formulated as independent single-task learning problems, with separate models optimized for each objective. Although such approaches can achieve strong task-specific performance, they do not explicitly exploit the complementary information shared between anatomically related tasks. Multi-task learning (MTL) addresses this limitation by jointly optimizing related objectives within a shared representation space, allowing information learned from one task to support the other. This shared learning process can also act as an implicit regularizer by discouraging the network from overfitting to task-specific noise and encouraging more generalizable feature representations \cite{cheng2022fully}.

The benefits of MTL have been demonstrated across several medical imaging applications. Xiao et al. \cite{xiao2023ctan}, for example, introduced a Cross-Task Attention Network that explicitly models interactions between multiple prediction tasks and showed that cross-task feature exchange can improve task-specific representations. Similarly, multi-scale MTL architectures have shown that jointly learning spatially related objectives can improve the utilization of features across different levels of the network \cite{bui2024multiscale}. Zhao et al. \cite{zhao2024yolomed} further demonstrated that multi-task feature interaction can integrate information across scales while maintaining an efficient shared inference framework.

This relationship is particularly relevant to bladder cancer MRI because tumor localization and assessment of muscle invasion are anatomically linked. Tumor segmentation requires accurate delineation of tumor extent and its boundary with surrounding bladder tissues, whereas MIBC assessment depends strongly on identifying whether the tumor disrupts or extends into the muscular layer of the bladder wall. The VI-RADS framework introduced by Panebianco et al. \cite{panebianco2025multiparametric} specifically evaluates MRI characteristics associated with muscular invasion, highlighting the importance of the spatial relationship between the tumor and bladder wall. Subsequent clinical validation by Wang et al. \cite{wang2019multiparametric} demonstrated that VI-RADS-derived MRI features can effectively discriminate muscle-invasive from non-muscle-invasive bladder cancer. Consequently, spatial features describing tumor location, extent, boundary characteristics, and interaction with the bladder wall are potentially informative for both segmentation and staging.

In our framework, this relationship is exploited through a shared multi-task representation and segmentation-guided feature interaction. The segmentation branch learns spatially precise tumor-related features, while the classification branch uses these features together with global anatomical context to predict MIBC status. Rather than treating segmentation and classification as two isolated prediction problems, the proposed formulation allows tumor localization to guide the staging decision. This design is particularly suitable for federated multi-center MRI, where learning task-shared and anatomically meaningful representations may improve robustness to scanner- and site-specific variations while reducing reliance on features that are specific to an individual institution.

\section{Methodology}
\label{sec:methodology}

\begin{figure}[t]
    \centering
    \includegraphics[width=\linewidth]{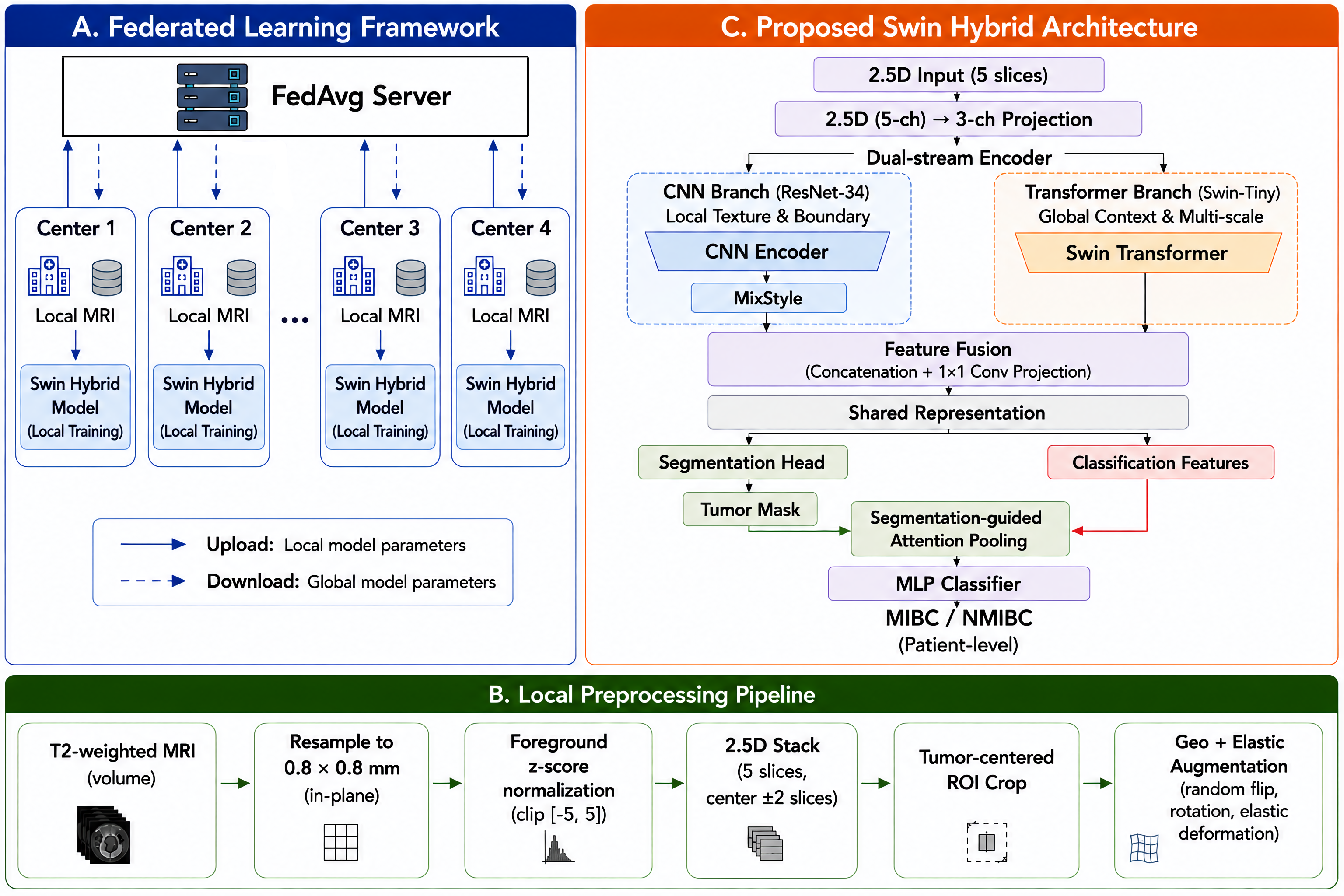}
    \caption{Proposed federated multi-task framework with FedAvg aggregation
    and the local Swin Hybrid network for tumor segmentation and MIBC
    classification.}
    \label{fig:framework}
\end{figure}

We formulate joint bladder tumor segmentation and muscle-invasive bladder
cancer (MIBC) classification as a federated multi-task learning problem, as
illustrated in Fig.~\ref{fig:framework}. Each hospital preprocesses its local
T2-weighted MRI scans into 2.5D inputs, applies the selected augmentation
strategy, and trains the proposed Swin Hybrid network. Only updated model
parameters are transmitted to the central server and aggregated using FedAvg
\cite{fedavg}, while patient data remain within their originating institutions.

Patient MRI data are distributed across $K=4$ hospitals. The federated dataset
is represented as

\begin{equation}
\mathcal{D}=\{\mathcal{D}_k\}_{k=1}^{K}, \qquad
\mathcal{D}_k=\{(x_i^k,y_i^k,c_i^k)\}_{i=1}^{n_k},
\label{eq:dataset}
\end{equation}

where $\mathcal{D}_k$ denotes the private dataset of hospital $k$. Each sample
contains a five-channel 2.5D MRI input
$x_i^k\in\mathbb{R}^{5\times H\times W}$, a tumor segmentation mask
$y_i^k\in\{0,1\}^{H\times W}$, and a binary MIBC/NMIBC label
$c_i^k\in\{0,1\}$.

The objective is to learn a shared model
$f_{\theta}(x)=(\hat{y},\hat{c})$ that jointly predicts the segmentation mask
and classification label by minimizing the sample-size-weighted loss

\begin{equation}
\theta^{*}
=
\arg\min_{\theta}
\sum_{k=1}^{K}
\frac{n_k}{N}\,
\mathcal{L}(\theta;\mathcal{D}_k),
\qquad
N=\sum_{k=1}^{K}n_k.
\label{eq:objective}
\end{equation}

The distributed objective is optimized using the federated training procedure
described below.
\subsection{Federated Learning Framework}

At communication round $t$, the server broadcasts the global model parameters
$\theta_t$ to all $K$ hospitals. Each hospital initializes its local model as
$\theta_{t,0}^{(k)}=\theta_t$ and performs $E=3$ local training epochs:

\begin{equation}
\theta_{t,e+1}^{(k)}
=
\theta_{t,e}^{(k)}
-
\eta
\nabla
\mathcal{L}\!\left(
\theta_{t,e}^{(k)};\mathcal{D}_k
\right),
\qquad e=0,\ldots,E-1,
\label{eq:localupdate}
\end{equation}

where $\eta$ denotes the local learning rate. After local training, hospital
$k$ sends the updated parameters
$\theta_t^{(k)}=\theta_{t,E}^{(k)}$ to the server. The server computes the
sample-size-weighted global model using FedAvg:

\begin{equation}
\theta_{t+1}
=
\sum_{k=1}^{K}
\frac{n_k}{N}
\theta_t^{(k)}.
\label{eq:fedavg}
\end{equation}

The aggregated model $\theta_{t+1}$ is redistributed to all hospitals, and the
procedure is repeated for $T=50$ communication rounds. All four hospitals
participate in every round and train the same multi-task Swin Hybrid model.

\subsection{Data Pre-processing and Harmonization}

To ensure the proposed federated multi-task framework remains resilient to the extensive variability introduced by multi-center data collection, a rigorous and standardized data preprocessing pipeline was deployed across all participating hospital clients. Multi-institutional T2-weighted MRI scans inherently exhibit profound intensity variations. These discrepancies arise from differing magnetic field strengths (e.g., 1.5 Tesla versus 3.0 Tesla scanners), varying manufacturer hardware, and diverse clinical acquisition protocols, which together introduce significant non-IID characteristics into the federated dataset.

To mitigate these domain shifts at the input level before model training, our pipeline first applies the N4 bias field correction algorithm to all raw MRI volumes. This step effectively removes low-frequency intensity non-uniformities caused by magnetic field inhomogeneities during image acquisition. Subsequently, to handle extreme intensity outliers and scanner-specific artifacts, we perform slice-wise intensity clipping, truncating voxel values at the 1st and 99th percentiles. Following outlier removal, the intensity distribution of each scan is normalized to a zero mean and unit variance using Z-score normalization.

To capture essential three-dimensional spatial context while circumventing the heavy computational burden and memory requirements of full 3D convolutional networks, we utilized a 2.5D input extraction strategy. For every target slice containing the bladder, we stack two adjacent slices from above and two from below along the z-axis. This process yields a robust five-channel input tensor $x_i^k \in \mathbb{R}^{5 \times H \times W}$ for the model. Furthermore, to ensure spatial consistency across the varying fields of view utilized by different clinical centers, all slices are resampled to a uniform in-plane spatial resolution of 0.5 mm $\times$ 0.5 mm using B-spline interpolation. Finally, the slices are either center-cropped or zero-padded to achieve a fixed input dimension of $256 \times 256$ pixels. By standardizing the physical spacing and intensity distributions locally at each client node, the Swin Hybrid network can focus entirely on learning clinically relevant anatomical features rather than compensating for raw scanner variance.

\subsection{Multi-task Hybrid CNN--Transformer Model}

Each client trains the proposed Swin Hybrid network for joint tumor segmentation
and MIBC classification. A lightweight adapter $\phi$ projects the five-channel
2.5D input into a three-channel representation,
$x_3=\phi(x)$, enabling the use of ImageNet-pretrained encoders.

\noindent\textbf{Dual-stream encoder.}
The encoder combines a pretrained ResNet-34 \cite{he2016deep} with a
Swin-Tiny Transformer \cite{liu2021swin}. The CNN branch extracts local texture
and boundary features $\{l_1,l_2\}$, while the Transformer branch captures
multi-scale global contextual features $\{s_1,s_2,s_3\}$. MixStyle
\cite{mixstyle} is applied to the CNN branch to improve robustness to
scanner-dependent appearance variations.

\noindent\textbf{Feature fusion.}
Intermediate CNN and Transformer features are fused as

\begin{equation}
z_{\mathrm{mid}}=\psi([s_1;l_2]),
\label{eq:fusion}
\end{equation}

where $\psi$ denotes channel projection, normalization, and nonlinear
activation. The resulting features
$\{s_3,s_2,z_{\mathrm{mid}},l_1\}$ are shared by the task-specific heads.

\noindent\textbf{Task-specific heads.}
An attention-gated decoder \cite{oktay2018attention} with deep supervision
produces the tumor mask $\hat{y}$. The predicted segmentation probability map
is used to guide spatial pooling over the deepest Transformer feature:

\begin{equation}
v=
\sum_{h,w}
s_3(h,w)
\frac{\sigma(\hat{y})(h,w)}
{\sum_{h',w'}\sigma(\hat{y})(h',w')}.
\label{eq:attnpool}
\end{equation}

The pooled representation $v$ is passed to a lightweight classifier to predict
the MIBC/NMIBC label $\hat{c}$. The segmentation map is detached before
attention pooling, preventing classification gradients from propagating through
the segmentation prediction.

\noindent\textbf{Joint objective.}
The local model is optimized using

\begin{equation}
\mathcal{L}
=
0.80\,\mathcal{L}_{\mathrm{seg}}
+
0.20\,\mathcal{L}_{\mathrm{cls}},
\label{eq:jointloss}
\end{equation}

where $\mathcal{L}_{\mathrm{seg}}$ combines Dice, binary cross-entropy,
Focal Tversky, and boundary losses with auxiliary deep-supervision terms.
The classification objective uses focal cross-entropy and is evaluated only
for tumor-bearing slices. Its weight is gradually increased during the first
20 cumulative training epochs.

\subsection{Augmentation Strategy}

To improve robustness to inter-center variations, each client applies the
Geo+Elastic augmentation strategy during local training. It combines random
horizontal flipping ($p=0.5$), rotation within $\pm15^\circ$ ($p=0.70$), and
elastic deformation ($\sigma=5.0$, $\alpha=18.0$, $p=0.20$). The geometric
transformations account for variations in patient positioning, while elastic
deformation \cite{simard2003best} models non-rigid variations in bladder and
tumor morphology.




\section{Experiments}

\subsection{Dataset and Data Partitioning}

We evaluate the proposed framework using the FedBCa dataset
\cite{cao2024multicenter}, which contains 275 T2-weighted MRI scans acquired
from four medical centers. Each scan is associated with a pixel-level bladder
tumor segmentation mask and a pathological MIBC/NMIBC label. Variations in
scanner hardware and acquisition protocols across centers introduce natural
data heterogeneity, providing an appropriate setting for evaluating the proposed
federated framework. Table~\ref{tab:dataset_stats} summarizes the center-wise
distribution of the dataset.

\begin{table}[t]
\caption{Center-wise distribution of the FedBCa dataset.}
\label{tab:dataset_stats}
\centering
\small
\begin{tabular}{lccc}
\toprule
\textbf{Center} & \textbf{Scans} & \textbf{NMIBC} & \textbf{MIBC} \\
\midrule
Center 1 & 160 & 130 & 30 \\
Center 2 & 48  & 23  & 25 \\
Center 3 & 32  & 19  & 13 \\
Center 4 & 35  & 18  & 17 \\
\midrule
Total    & 275 & 190 & 85 \\
\bottomrule
\end{tabular}
\end{table}

A patient-level stratified split was performed independently within each
center, preserving the MIBC/NMIBC class distribution as closely as possible.
Approximately 71\% of the patients were assigned to training, 10\% to
validation, and 19\% to testing. The same center-wise partitions were used
for both centralized and federated experiments, ensuring that no patient
contributed data to more than one split.

\subsection{Implementation and Training Details}

All experiments were implemented in PyTorch ($\geq$2.1) and conducted on an
NVIDIA GeForce RTX 4070 GPU (12\,GB VRAM). ImageNet-pretrained
backbones were obtained using \texttt{timm} (v0.9.16), while federated
training was coordinated using Flower \cite{beutel2007flower}. Models were optimized using AdamW
\cite{loshchilov2017decoupled} with a warmup-cosine learning-rate schedule
and gradient clipping with a maximum norm of 1.0.

For the proposed Swin Hybrid model, the base learning rate was set to
$2\times10^{-5}$, with learning-rate multipliers of 0.15 and 0.30 for the
ResNet-34 and Swin-Tiny branches, respectively. A weight decay of
$1\times10^{-4}$ and an effective batch size of 24 were used. The
classification loss was gradually introduced during the first 20 cumulative
training epochs, as described in Section~\ref{sec:methodology}.

For federated training, each medical center was treated as an independent
client. All four clients participated in every communication round, with
$E=3$ local epochs per round and $T=50$ communication rounds. Client
parameters were aggregated using sample-size-weighted FedAvg
(Eq.~\ref{eq:fedavg}), and the global model with the highest validation
combined score was retained for final evaluation.

\subsection{Evaluation Metrics}

Tumor segmentation performance was evaluated using the Dice Similarity
Coefficient (DSC),

\begin{equation}
\mathrm{DSC}
=
\frac{2|P\cap G|+\epsilon}
{|P|+|G|+\epsilon},
\qquad
\epsilon=10^{-5},
\label{eq:dsc}
\end{equation}

where $P$ and $G$ denote the predicted and ground-truth tumor masks,
respectively. DSC was computed only on slices containing a non-empty
ground-truth tumor mask.

Classification performance was evaluated using the area under the receiver
operating characteristic curve (AUC) at the patient level. For each patient,
slice-level MIBC probabilities were averaged to obtain a single patient-level
prediction score.

For checkpoint and model selection, a combined validation score was defined as

\begin{equation}
S_{\mathrm{comb}}
=
0.55\,\mathrm{DSC}
+
0.45\,\mathrm{AUC}.
\label{eq:combined}
\end{equation}

The combined score was used only for model selection, while DSC and AUC were
reported separately as the primary segmentation and classification metrics.

\begin{table}[t]
\caption{Centralized performance comparison of the candidate architectures.
Best results are shown in bold.}
\label{tab:arch_comp}
\centering
\small
\begin{tabular}{lccc}
\toprule
\textbf{Single-Task Models} & \textbf{DSC} & \textbf{AUC} & \textbf{Combined} \\
\midrule
\textbf{ResNet50 U-Net}              & \textbf{0.8181} & \textbf{0.8304} & \textbf{0.8236} \\
EfficientNet-B0 U-Net       & 0.7669 & 0.7589 & 0.7633 \\
\midrule
\textbf{Multi-Task Models} & \textbf{DSC} & \textbf{AUC} & \textbf{Combined} \\
\midrule
ResNet50+CBAM     & 0.7579 & 0.7232 & 0.7423 \\
Multi-task EfficientNet-B0  & 0.7653 & 0.7679 & 0.7664 \\
\textbf{Swin Hybrid}        & \textbf{0.8076} & \textbf{0.8214} & \textbf{0.8138} \\
\bottomrule
\end{tabular}
\end{table}

\subsection{Experimental Protocol}

The experimental evaluation was conducted in two stages. First, five candidate architectures were compared under centralized training: ResNet50 U-Net \cite{he2016deep,ronneberger2015u}, EfficientNet-B0 U-Net \cite{tan2019efficientnet,ronneberger2015u}, ResNet50+CBAM \cite{woo2018cbam}, multi-task EfficientNet-B0 \cite{tan2019efficientnet}, and the proposed Swin Hybrid model, which combines a ResNet-34 backbone \cite{he2016deep} with a Swin-Tiny Transformer \cite{liu2021swin}. All architectures were evaluated using the same patient-level data partitions and evaluation metrics, and the model with the highest combined validation score was selected for the subsequent augmentation study.

Using the selected Swin Hybrid model, we then evaluated seven augmentation configurations: No Augmentation, GIN Only, Augmentation Stack, Geo+Elastic, Geo+Bias, Geo+Gamma, and Geo+Noise. GIN follows the nonlinear intensity augmentation strategy introduced in FedGIN \cite{nagaraju2025fedgin}. The augmentation study was performed under centralized training and then repeated under federated training using FedAvg across all four participating centers. The same architecture, patient-level data partitions, and evaluation protocol were maintained throughout, enabling a direct comparison of the augmentation strategies under both training settings.

\section{Results}

\subsection{Centralized Architecture Comparison}

Table~\ref{tab:arch_comp} compares the candidate architectures under centralized training. Among the single-task models, ResNet50 U-Net achieved the highest segmentation accuracy (DSC of 0.5708). The proposed multi-task Swin Hybrid provided the best overall balance between tasks, achieving the highest patient-level AUC (0.8214) and combined score (0.8138). Consequently, the Swin Hybrid was selected as the optimal architecture for the subsequent augmentation experiments.

\subsection{Centralized and Federated Augmentation Comparison}

Table~\ref{tab:aug_comparison} compares the seven augmentation strategies
using the selected Swin Hybrid model under centralized and federated training.

Under centralized training, the Augmentation Stack achieved the highest DSC (0.8487), whereas "No Augmentation" setup achieved the highest AUC (0.9286) and Geo+Elastic had the highest combined score (0.8688). Although the Augmentation Stack produced the best segmentation accuracy, Geo+Elastic provided the best balance between segmentation and patient-level classification, making it the strongest augmentation strategy.

Under federated training, Geo+Elastic again achieved the highest AUC (0.8931) and combined score (0.8474), with a DSC of 0.8100. Although the Augmentation Stack produced the highest DSC (0.8293), its lower AUC resulted in a lower combined score. Therefore, Geo+Elastic was selected as the augmentation strategy for the final federated Swin Hybrid model.

\begin{table}[t]
\caption{Centralized and federated augmentation comparison using the
Swin Hybrid model. Best results within each training setting are shown in bold.}
\label{tab:aug_comparison}
\centering
\small
\setlength{\tabcolsep}{3.5pt}
\begin{tabular}{lcccccc}
\toprule
& \multicolumn{3}{c}{\textbf{Centralized}}
& \multicolumn{3}{c}{\textbf{Federated}} \\
\cmidrule(lr){2-4}
\cmidrule(lr){5-7}
\textbf{Augmentation}
& \textbf{DSC}
& \textbf{AUC}
& \textbf{Comb.}
& \textbf{DSC}
& \textbf{AUC}
& \textbf{Comb.} \\
\midrule
None
& 0.8076 & 0.8214 & 0.8138
& 0.7908 & 0.7716 & 0.7822 \\

GIN Only
& 0.8161 & 0.8750 & 0.8426
& 0.7790 & 0.7270 & 0.7560 \\

Aug. Stack
& \textbf{0.8487} & 0.8661 & 0.8565
& \textbf{0.8220} & 0.8230 & 0.8220 \\

\textbf{Geo+Elastic}
    & 0.8272 & \textbf{0.9196 }& \textbf{0.8688}
& 0.8100 & \textbf{0.8931} & \textbf{0.8474} \\

Geo+Bias
& 0.8391 & 0.8661 & 0.8513
& 0.8094 & 0.8414 & 0.8238 \\

Geo+Gamma
& 0.8417 & 0.8750 & 0.8567
& 0.7987 & 0.8575 & 0.825 \\

Geo+Noise
& 0.8398 & 0.8750 & 0.8556
& 0.8108 & 0.8240 & 0.8167 \\
\bottomrule
\end{tabular}
\end{table}

Overall, Geo+Elastic consistently achieved the highest combined performance under both centralized and federated training, with combined scores of 0.8688 and 0.8474, respectively. Although federated training showed modest reductions in DSC (0.8272 to 0.8100) and AUC (0.9196 to 0.8931), Geo+Elastic remained the most effective augmentation strategy across both training settings.

\subsection{Discussion and Impact of Multi-Task Learning}

Our experimental evaluation underscores the advantages of federated multi-task learning (MTL) over conventional single-task architectures. While single-task baselines like the ResNet50 U-Net achieved a high Dice Similarity Coefficient (DSC) for dense pixel-level predictions, they fundamentally lacked the global contextual awareness needed to maximize patient-level classification accuracy (AUC).

This reveals a critical trade-off in medical imaging: high-capacity CNNs excel at extracting localized boundaries but often struggle with the long-range spatial dependencies vital for cancer staging. The proposed Swin Hybrid architecture elegantly bridges this gap by delegating local feature extraction to the ResNet-34 branch and global anatomical modeling to the Swin-Tiny Transformer. Furthermore, our attention-gated spatial pooling mechanism restricts the classification head's receptive field to predicted tumor regions, actively preventing the model from hallucinating MIBC predictions based on irrelevant background noise or benign variations in the bladder wall.

Finally, our augmentation analysis provided insights into handling multi-center heterogeneity. While the Augmentation Stack yielded the highest centralized DSC (0.8487), the Geo+Elastic strategy proved optimally balanced for federated settings, achieving the highest combined score (0.8474) and preserving an impressive AUC (0.8931). By simulating natural, non-rigid biomechanical variations of the bladder volume and irregular tumor morphologies, elastic deformation acted synergistically with the MTL objective as a powerful regularizer against non-IID client drift. Ultimately, these results validate that combining multi-task feature sharing with targeted elastic data augmentation allows federated models to achieve centralized-level robustness without compromising patient privacy.

\section{Conclusion}

This work introduced a federated multi-task learning framework for joint bladder tumor segmentation and MIBC classification from multi-center T2-weighted MRI. The proposed Swin Hybrid model combined CNN-based local feature extraction with Transformer-based global contextual modeling, achieving a strong balance between segmentation and classification performance. The augmentation analysis further showed that Geo+Elastic augmentation was the most effective strategy across centralized and federated settings, particularly under heterogeneous multi-center data. Overall, the results demonstrate that the proposed framework can support collaborative bladder cancer analysis across institutions without centralizing patient data, while maintaining competitive performance on both clinical tasks.

\bibliographystyle{splncs04}%
\bibliography{ref}

\end{document}